\documentclass{article}
\usepackage{spconf,amsmath,graphicx}
\usepackage{adjustbox}
\usepackage{color}
\usepackage{xcolor}
\usepackage{multirow}
\usepackage{multicol}
\usepackage{caption}
\usepackage{subcaption}
\usepackage{url}
\usepackage{hyperref}

\def\arrvline{\hfil\kern\arraycolsep\vline\kern-\arraycolsep\hfilneg}
\definecolor{weining}{HTML}{246FFF}
\definecolor{maryam}{HTML}{12b637}
\definecolor{todo}{HTML}{E74C3C}

\newcommand{\hubert}{HuBERT}

\title{Cocktail HuBERT: Generalized Self-Supervised Pre-training \\for Mixture and Single-Source Speech}
\name{Maryam Fazel-Zarandi and Wei-Ning Hsu}
\address{
    Meta AI - FAIR\\
    \texttt{\{maryamfazel,wnhsu\}@meta.com}
}

\begin{document}
\ninept
\maketitle
\begin{abstract}
Self-supervised learning leverages unlabeled data effectively, improving label efficiency and generalization to domains without labeled data. While recent work has studied generalization to more acoustic/linguistic domains, languages, and modalities, these investigations are limited to single-source speech with one primary speaker in the recording. This paper presents Cocktail HuBERT, a self-supervised learning framework that generalizes to mixture speech using a masked pseudo source separation objective. This objective encourages the model to identify the number of sources, separate and understand the context, and infer the content of masked regions represented as discovered units. Cocktail HuBERT outperforms state-of-the-art results with $69\%$ lower WER on multi-speaker ASR, $31\%$ lower DER on diarization, and is competitive on single- and multi-speaker tasks from SUPERB. 

\end{abstract}
\section{Introduction}
Self-supervised learning (SSL) has greatly advanced speech processing over the past few years~\cite{chung2019unsupervised, schneider2019wav2vec, Hsu2021HuBERTSS, baevski2022data2vec, chen2022wavlm}. Supervised fine-tuning from a pre-trained model enjoys better label efficiency, achieving performance on par with supervised models using hundredths fewer labeled data~\cite{Baevski2020}. The representations learned with pre-trained models are more universal: in contrast to those from supervised learning~\cite{chen2021speechnet, hsu2019transfer, jia2018transfer}, self-supervised representations benefit a wider range of tasks~\cite{Yang2021, Tsai2022}. Self-supervised representations also enable many novel applications, such as unsupervised speech recognition and synthesis~\cite{baevski2021unsupervised, liu2022simple, ni2022unsupervised}, disentangled speech codec~\cite{polyak2021speech}, text-free spoken language models and prompting~\cite{lakhotia2021generative, chang2022exploration}. 

A key advantage of self-supervised pre-training is that it uses unlabeled data instead of labeled data, such that a model can be pre-trained on data covering more domains~\cite{hsu2021robust, kawakami2020learning, conneau2020unsupervised, babu2021xls}. Consequently, the fine-tuned model is more robust to domain shift, suffering milder degradation when evaluated on domains unseen during fine-tuning~\cite{hsu2021robust}. Generalizing this idea, recent work also extends self-supervised pre-training to multi-modal speech~\cite{shi2022learning,hsu2022single} and demonstrates a multi-modal speech recognition system can be built with only labeled unimodal data. 
However, up until now, speech pre-training has been designed for single-source speech, which contains one primary speaker in each sample whereas other sources are assumed noise~\cite{chen2022wavlm}, leaving mixture speech unattended.

Mixture speech, where multiple speakers may speak at the same time, occurs frequently in conversational scenarios.
These scenarios impose greater challenges to applications common to single-source speech (e.g., recognizing the ``target'' speech from a mixture), and also generate applications specific to mixture speech, including speech diarization, source separation, multi-speaker speech recognition (transcribe everything) and more. Pre-trained models designed for single-source speech are likely to be sub-optimal for these tasks.

In an effort to broaden the applicability of pre-trained models to wider speech varieties, this paper presents Cocktail HuBERT (C-HuBERT), a self-supervised framework that pre-trains on both single-source and mixture speech with a unified objective. The pre-training objective can be summed as \textit{masked pseudo source separation}, which predicts automatically discovered units of randomly masked spans for each source in the mixture given unmasked speech context. The speech mixture contains one or more sources, created by artificially mixing single source samples. To excel at this task, the model is required to perform three tasks jointly: source separation, acoustic modeling, and language modeling.
Evaluation on multi-speaker automatic speech recognition (MS-ASR) and speech diarization (SD) verifies that the proposed objective is particularly effective for downstream tasks concerning mixture speech, outperforming state-of-the-art results by large margins (7.8\% vs 24.9\% WER on Libri2Mix MS-ASR, and 3.86\% vs 5.62\% DER on Libri(2+3)Mix SD). 
Evaluation on SUPERB~\cite{Yang2021} also shows strong performance on additional mixture tasks and slight-to-none degradation on single-source tasks.

\begin{figure*}[t]       
    \begin{subfigure}[t]{0.24\linewidth}
        \includegraphics[trim={220 50 330 40},clip,width=\linewidth]{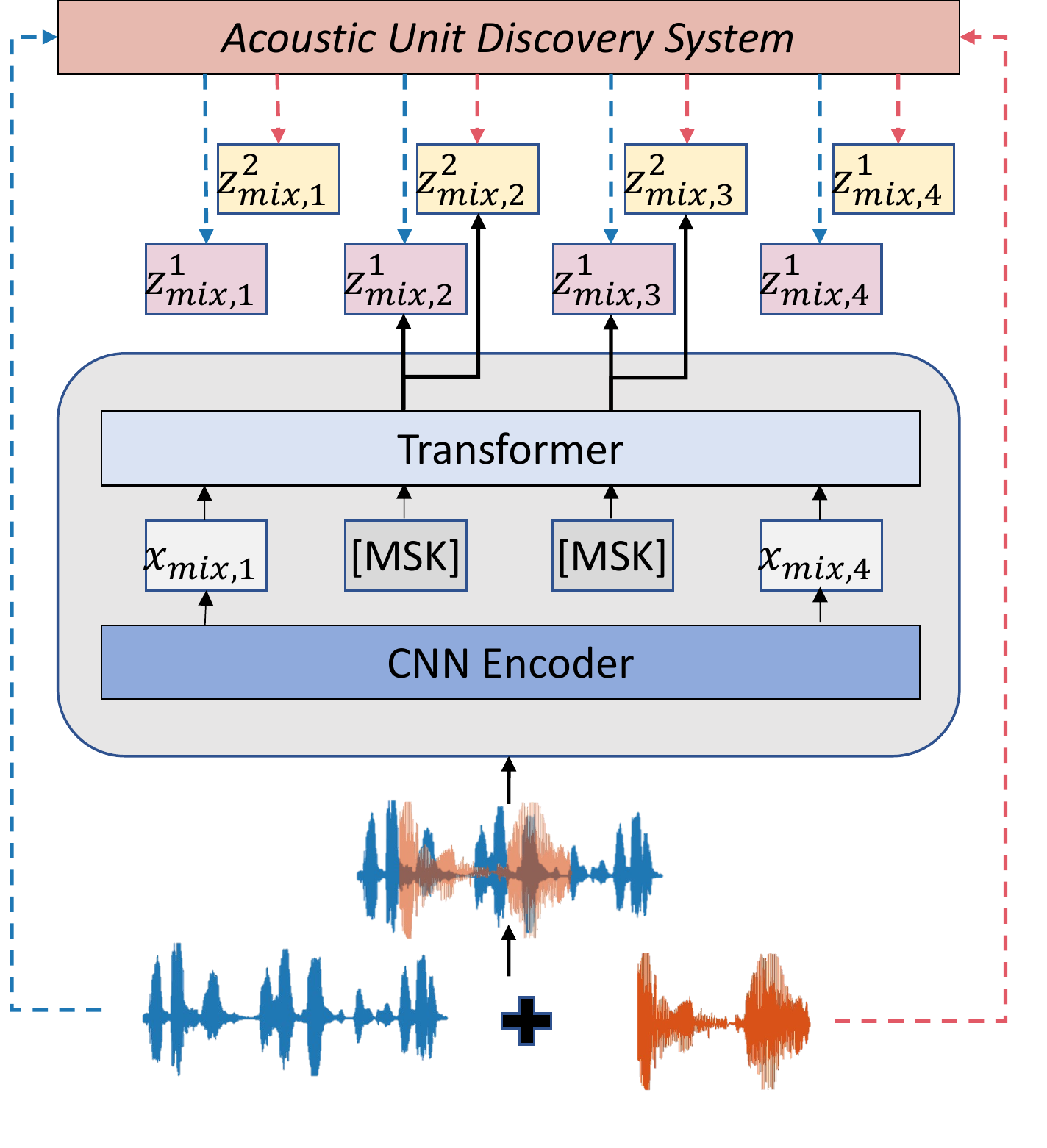}   
        \caption{Cocktail HuBERT}
        \label{fig:model}
    \end{subfigure}
    \hfill
    \begin{subfigure}[t]{0.74\linewidth}
        \includegraphics[trim={0 230 0 0},clip,width=\linewidth]{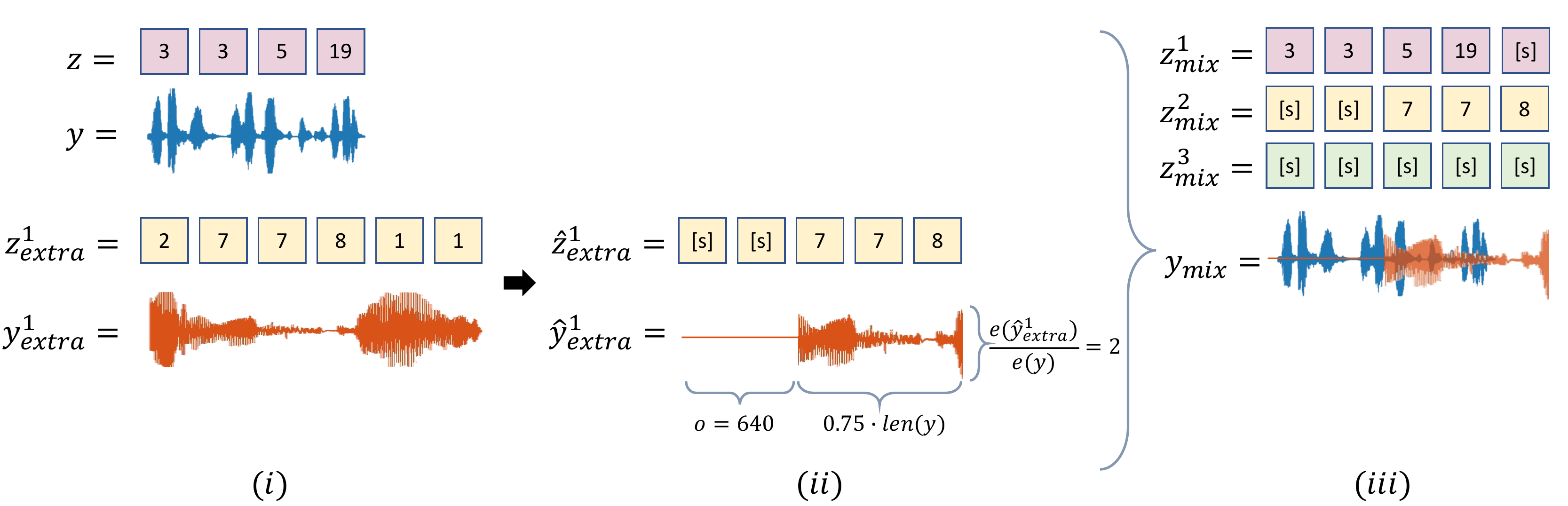}   
        \caption{Mixture simulation.}
        \label{fig:mix}
    \end{subfigure}
    \vspace{-5pt}
    \caption{(a) C-HuBERT ($K=2$) predicts hidden units of the masked frames for each source in the input audio mix generated by k-means clustering. (b) Mixture simulation with $K=3$, $n=1$, $(r_l, r_e, o) = (0.75, 2.0, 640)$. Step i: sample the number of extra sources $n$ and then sample additional sources $z_{extra}^{1:n}$. Step ii: chunk, scale, and shift according to sampled $(r_l, r_e, o)$. $e(y)$ denotes the energy of $y$. Step iii: mix audio and pad target units with [SIL] for silent frames (last frame of $z_{mix}^1$ and first two frames of $z_{mix}^2$) and silent streams ($z_{mix}^3$).} 
    \vspace{-12pt}
\end{figure*}

\section{Background}
This paper is built upon Hidden Unit BERT (\hubert{})~\cite{Hsu2021HuBERTSS}, one of the state-of-the-art speech pre-training frameworks. 
The pre-training objective of \hubert{} is masked cluster prediction. Similar to BERT, it masks part of the speech input, and predicts given the context (unmasked part) some label derived from the masked input. While the label used in BERT is the input token itself, \hubert{} proposes to obtain discrete labels via clustering audio features and refine the label iteratively. 
Concretely, let $y$ be waveform, $x_t = f_t(y)$ be \hubert{} local feature extractor $f$ (CNN) output at time $t$, $c^l_t = g^l_t(x)$ be contextualized feature extractor $g$ ($L$-layer Transformer) output at time $t$ layer $l$, and $z_t$ be the target unit at time $t$. \hubert{} ($f$ and $g$) is pre-trained by predicting $z_t$ from $g^L_t(\text{MASK}(x))$ for time steps $t$ that are masked, where $\text{MASK}(\cdot)$ is an operation that randomly samples spans of 10 frames and replaces the features in those spans with a learned masked embedding $\text{[MSK]}$ following wav2vec2.0~\cite{Baevski2020}.
In the first iteration, $z_t$ are obtained by clustering MFCC features. In the subsequent iterations, the latest iteration \hubert{} representations $c^l_t$ are used for clustering, which produces higher quality cluster assignments than those from raw or earlier iteration \hubert{} features.

Intuitively, \hubert{} pre-training solves acoustic and language modeling task jointly, where the model needs to understand the content from observed regions (acoustic model) and then infer the label for the masked frames (language model).

\section{Cocktail HuBERT}
\label{cocktail-hubert}
The ``cocktail party problem'' describes the setup where sounds from different sources are mixed prior to being perceived, such that estimation of individual sources given the mixture is required to perform downstream tasks. Human brains have impressive ability to focus on a particular stimulus leveraging structural properties of single sources. Researchers have also attempted to reproduce this capability and develop applications like source separation~\cite{shi2021discretization} and multi-speaker speech recognition~\cite{Chang2022}. 

In this paper, we present Cocktail HuBERT, a self-supervised learning objective that trains the model to tackle the cocktail party problem, as depicted in Figure~\ref{fig:model}. Conceptually, the model takes as input a masked mixture which contains one or more speech sources with spans of frames randomly masked, similar to \hubert{}. The model processes the masked mixture and predicts the discrete labels for each source for only the masked frames. The discrete labels can be viewed as automatically discovered frame-level phonetic transcripts. Thus, the pre-training task is analogous to pseudo source separation that predicts a proxy frame representation for each source.

\subsection{Model and Training Objective}
Cocktail HuBERT adopts the same local and contextualized feature encoder ($f$ and $g$) as HuBERT, but instead of having only one projection head to predict the posterior over units for a single source, Cocktail HuBERT has $K$ project heads to predict units for at most $K$ sources. Let $y_{mix}$ be a mixture containing up to $K$ sources and $z^{i}_{mix}$ for $i \in [K]$ be the target unit sequence for source $i$ (some corresponds to silent sources). The model outputs $K$ streams of predictions, where $p^j_t(\cdot \mid g(\text{MASK}(f(y_{mix})))$ denotes the $j$-th stream prediction at time step $t$. The loss of the $j$-th stream prediction with respect to the $i$-th source unit sequence $z^i_{mix}$ is:

\vspace{-5pt}
\begin{equation}
    L_m^{j,i} = \sum_{t \in M} \log p^j(z^i_{mix,t} \mid g(\text{MASK}(f(y_{mix}))),
\end{equation}
where $M$ denotes the masked time steps. Since the order of the model predictions $j$ do not necessarily align with the order of the sources $i$, permutation invariant training (PIT) loss~\cite{Yu2017} is deployed which finds the alignment with the minimum loss. Let $\cal P$ be all permutations of $[K]$, the masked pseudo source separation objective is
\vspace{-5pt}
\begin{equation}
    \frac{1}{K} \arg \min_{\pi \in {\cal P}} \sum^{K}_{j = 1} L_m^{j, \pi(j)}.
\end{equation}
\vspace{-10pt}

\subsection{Mixture Simulation}
\label{data-mixing}

For training our models, overlapped speech for a maximum of $K$ speakers is simulated as follows (See Fig~\ref{fig:mix}).
A batch of $B$ utterances where $B \ge K$ is sampled from the dataset. 
For each utterance $y$ and its units $z$, the number of additional sources $n \in \{0, \cdots K-1\}$ is sampled with $P(n=0) = 1 - p_{mix}$ and $ P(n=k) = p_{mix} / (K-1)$ for $\forall k \ne 0$.
An additional source is either an utterance from the batch, or a non-speech noise sampled from a noise dataset. The probability of selecting noise is $p_{noise}$.
For each additional source $y_{extra}^{k}$, a tuple of length ratio, energy ratio, and offset $(r_l, r_e, o)$ are sampled from some uniform distributions, which are used to chunk, scale, and shift $y_{extra}^{k}$ with respect to $y$. Let $\hat{y}_{extra}^{k}$ be the resulting source and $\hat{z}_{extra}^k$ be the units chunked correspondingly if $y_{extra}^{k}$ is not noise. The resulting mixture $y_{mix} = y + \sum_{k=1}^{n} \hat{y}_{extra}^{k}$. 
Note that each source is right-padded to the maximum length among $y$ and $y_{extra}^{k}\; \forall k$ with silence. A special token [SIL] is used for frames corresponding to padded silence (including the offset at the beginning). The first $n+1$ unit sequences correspond to the [SIL]-padded $z$ and $\hat{z}_{extra}^k$ for non-noise $k$. The remaining $K - (n + 1)$ sequences and those corresponding to noise samples are set to [SIL] sequences.

\section{Related Work}
Cocktail HuBERT is most related to HuBERT~\cite{Hsu2021HuBERTSS} and WavLM \cite{chen2022wavlm}, which are self-supervised speech pre-training frameworks based on masked cluster prediction. Similar to Cocktail HuBERT, WavLM also stochastically mixes single source speech with noise and/or other single source speech. However, WavLM is effectively HuBERT with data augmentation, which pre-trains the model using the same objective as HuBERT where only the units of the ``primary'' speech are predicted: the added noise and speech are both treated as noise and should be ignored by the model. Consequently, for the model to differentiate primary speech from interfering speech, WavLM deploys a more restrictive mixing strategy, where the interfering audio is at most half as long as the primary speech. 

On the other hand, Cocktail HuBERT and \cite{shi2021discretization} also share similar training objectives. Inspired by the capability of converting units back to speech~\cite{polyak2021speech} and the connection between single/multi-speaker speech recognition to speech enhancement/separation, \cite{shi2021discretization} proposes an alternative approach to speech enhancement and source separation by predicting units instead of spectral masks or waveforms. Comparing Cocktail HuBERT with \cite{shi2021discretization}, the former are designed for pre-training, which masks partial input and predicts units only for masked frames, requiring the model to perform language modeling. It is also evaluated on a wide range of downstream tasks including both mixture and single-source speech processing. In contrast, the latter is designed for particular downstream tasks (enhancement and separation) that predict units for all frames without masking the input. It is unclear how the resulting model performs on other tasks.


\section{Experimental Setup}
\label{tasks}

For unsupervised pre-training, we use 960 hours of LibriSpeech audio \cite{LibriSpeech} for the \textsc{Base} model and 60k hours of Libri-light audio \cite{LibriLight} for the \textsc{Large} model. 
We extract features from the $9$-th transformer layer of the HuBERT \textsc{Base} model for K-means with $500$ clusters and use those labels to train the Cocktail HuBERT models.
This ensures that we have high quality labels. 
We apply data mixing to $20\%$, $60\%$, and $100\%$ of the data, where the mixing noise probability is set to $10\%$.
The Cocktail \textsc{Base} and \textsc{Large} models are trained on $32$ and $128$ GPUs, respectively, for $400$k and $800$k steps. The batch sizes are at most $87.5$ and $56.25$ seconds of audio per GPU.
Adam \cite{Kingma2014} optimizer is used with $\beta = (0.9,0.98)$, and learning rate ramps up linearly from $0$ to peak learning for the first $32$k training steps, and then decays back to zero. The peak learning rates are 5e-4/1e-4 for Cocktail \textsc{Base}/\textsc{Large} models. 
$K/p_{mix}$ are set to 5/1.0 for \textsc{Base} and 3/1.0 for \textsc{Large} unless otherwise specified.

We evaluate our pre-trained models specifically on multi-speaker speech recognition and diarization tasks, as they involve multiple speakers. 
We use the LibriMix data \cite{cosentino2020librimix} which contains multi-speaker overlapped speech simulated by mixing utterances from the LibriSpeech corpus. We focus on the two- and three-speaker scenarios, 
and mixes with the ``max mode'' where the shortest utterance is padded to the longest one.
For MS-ASR, we fine-tune our models on multi-speaker labeled data using the connectionist temporal classification (CTC) \cite{Graves2006} loss for each (output stream, label stream) pair and compute the PIT-CTC loss, to fine-tune the whole model except for the local feature extractor. The projection layers are removed and replaced with a randomly initialized softmax layer for each stream. The CTC target vocabulary includes 26 English characters, a space token, an apostrophe, and a special CTC blank symbol.
We fine-tune each model on 8 GPUs on the \texttt{train-100-mix-clean} subset of Libri2Mix for the 2-speaker scenario. 
The batch sizes per GPU are at most 200/80 seconds of audio for \textsc{Base}/\textsc{large} models. We sweep over peak learning rate ([1e-5, 1e-4]) for each model size using the PIT word error rate (WER) on the \texttt{dev\_mix\_clean} subset as criterion for model selection. All other hyperparameters are based on \cite{Hsu2021HuBERTSS}, except that we set \textit{freeze-step} to zero. We use beam search decoding with a 4-gram language model and a beam size of $500$.

For SD, we use a similar setup to SUPERB \cite{Yang2021}, where we freeze the pre-trained model and weight-sum the representations from different $g$ layers with learnable weights as the input to the diarization model. The diarization model uses a single layer 512-unit LSTM and is trained with the PIT loss.
We train each model on 1 GPU on the \texttt{train-100-mix-both} subset of Libri2Mix and Libri3Mix, and a mixture of both datasets. We use a batch size of 8, train for 30k steps, and sweep over learning rate ([1e-2, 1e-4]) for each model using accuracy on the \texttt{dev\_mix\_both} subset of each dataset as criterion for model selection. For the evaluation metric, we use the diarization error rate (DER) \cite{Fiscus2006}. 

We also compare our models to other strong pre-trained models on a subset of SUPERB tasks~\cite{Yang2021, Tsai2022}, including
Phoneme Recognition (PR), Automatic Speech Recognition (ASR), Keyword Spotting (KS), Query by Example Spoken Term Detection (QbE), Intent Classification (IC), Slot Filling (SF), Emotion Recognition (ER), Speech Enhancement (SE), and Speech Separation (SS) following the protocol. Overall score is computed following \cite{chen2022wavlm}. 

\section{Results}



\begin{table}[!b]
\vspace{-12pt}
\caption{WERs of multi-speaker ASR models trained on Libri2Mix and tested on \texttt{dev-mix-clean} (dev-mix) and \texttt{test-mix-clean} (test-mix). C-HuBERT \textsc{Base} $p_{mix}=0.6$.}
\label{ms-asr-table}
\vspace{-15pt}
\begin{center}
\begin{adjustbox}{width=0.85\columnwidth}
\begin{tabular}{l|c|cc}
\hline
\multirow{2}{*}{\bf Model} &\multirow{2}{*}{\bf LM} & \multicolumn{2}{c}{\bf Libri2Mix} \\
& & \multicolumn{1}{c}{\bf dev-mix} & \multicolumn{1}{c}{\bf test-mix} \\
 \hline
PIT-CTC \cite{Chang2022} & Transformer & $24.0$ & $26.3$ \\
GTC-e \cite{Chang2022} & Transformer & $32.7$ & $33.7$\\
Cond-Conformer-CTC \cite{ConditionalConformer} & greedy & $24.5$ & $24.9$ \\
\hline
HuBERT \textsc{Base} & 4-gram & $35.8$ & $37.2$ \\
HuBERT \textsc{Large} & 4-gram & $34.4$ & $35.2$\\
\hline
C-HuBERT \textsc{Base} & 4-gram & $12.7$ & $13.7$ \\
C-HuBERT \textsc{Large} & 4-gram & $\textbf{6.6}$ & $\textbf{7.8}$ \\
\hline
\end{tabular}
\end{adjustbox}
\end{center}
\vspace{-10pt}
\end{table}

\begin{table}[!b]
\vspace{-5pt}
\caption{WERs of multi-speaker and single-speaker ASR tested on single-speaker LibriSpeech sets: \texttt{\{dev,test\}-\{clean,other\}} (\{d,t\}-\{c,o\}). C-HuBERT \textsc{Base} $p_{mix}=0.6$. 4-gram LMs are used}
\label{ms-asr-single}
\vspace{-5pt}
\begin{adjustbox}{width=\columnwidth}
\begin{tabular}{l|cccc|cccc}
\hline
\multicolumn{1}{c|}{\bf Model}
    & \multicolumn{4}{c|}{\bf finetune on LS-10h} & \multicolumn{4}{c}{\bf finetune Libri2Mix} \\
    & \multicolumn{1}{c}{\bf d-c} & \multicolumn{1}{c}{\bf d-o} &\multicolumn{1}{c}{\bf t-c} &\multicolumn{1}{c|}{\bf t-o} 
    & \multicolumn{1}{c}{\bf d-c} & \multicolumn{1}{c}{\bf d-o} &\multicolumn{1}{c}{\bf t-c} &\multicolumn{1}{c}{\bf t-o} \\
 \hline
HuBERT \textsc{B} & $\textbf{4.1}$ & $\textbf{9.4}$ & $\textbf{4.5}$ & $\textbf{9.7}$ & $9.1$ & $20.7$ & $10.1$ & $23.6$ \\
C-HuBERT \textsc{B}  & $4.4$ & $10.3$ & $4.9$ & $10.9$ & $\textbf{4.9}$ & $\textbf{13.0}$ & $\textbf{6.3}$ & $\textbf{13.7}$ \\
\hline
HuBERT \textsc{L} & $\textbf{2.5}$ & $\textbf{5.2}$ & $\textbf{3.0}$ & $\textbf{5.6}$ & $5.8$ & $14.3$ & $8.1$ & $14.7$ \\
C-HuBERT \textsc{L}  & $2.8$ & $5.9$ & $3.3$ & $6.4$ & $\textbf{3.7}$ & $\textbf{9.6}$ & $\textbf{4.5}$ & $\textbf{10.5}$ \\
\hline
\end{tabular}
\end{adjustbox}
\end{table}

\subsection{Multi-speaker and Single-speaker ASR}

We first evaluate Cocktail HuBERT models on multi-speaker ASR and compare them to three state-of-the-art supervised baselines: (1) the end-to-end ASR model trained with PIT-CTC \cite{Chang2019}, (2) the end-to-end ASR model trained with the extended Graph-based temporal classification \cite{Chang2022} loss (GTC-e), and (3) the Conditional-Conformer-CTC model \cite{ConditionalConformer} that generates CTC predictions conditioning on past CTC predictions for other streams. To understand how pre-training objectives affect the multi-speaker ASR performance, we also fine-tune HuBERT \textsc{Base} and \textsc{Large}, which have the identical model architecture as C-HuBERT models, as our self-supervised baselines.

\begin{table*}[!t]
\caption{Universal Speech Representation Evaluation on SUPERB. Results from all other models are from \cite{chen2022wavlm}. \textsc{Base}/\textsc{Large} models are pre-trained on LS 960hr/Libri-light 60k hr, except for WavLM \textsc{Large}$\dagger$ which was trained on Mix 94k hr}
\label{superb}
\vspace{-5pt}
\begin{adjustbox}{width=\textwidth}
\begin{tabular}{l|c||c|c|c|c||c|cc||c||cc|c||c}
\hline
& & \multicolumn{4}{c||}{Content} & \multicolumn{3}{c||}{Semantics} & \multicolumn{1}{c||}{ParaL} & \multicolumn{3}{c||}{Generation} & Overall \\
\cline{3-14}
\multicolumn{1}{c|}{Method} &\multicolumn{1}{c||}{$\#$Params} & \multicolumn{1}{c|}{PR} & \multicolumn{1}{c|}{ASR} & \multicolumn{1}{c|}{KS} & \multicolumn{1}{c||}{QbE} & \multicolumn{1}{c|}{IC} & \multicolumn{2}{c||}{SF} & \multicolumn{1}{c||}{ER} & \multicolumn{2}{c|}{SE} & \multicolumn{1}{c||}{SS} & \\\cline{3-14}
& & PER $\downarrow$ & WER $\downarrow$ & Acc $\uparrow$ & MTWV $\uparrow$ & Acc $\uparrow$ & F1 $\uparrow$ & CER $\downarrow$ & Acc $\uparrow$ & PESQ $\uparrow$ & STOI $\uparrow$ & SI-SDRi $\uparrow$ & Score $\uparrow$ \\ 
\hline
wav2vec 2.0 \textsc{Base} \cite{Baevski2020} & 95.04M  &  5.74 & 6.43 & 96.23 & 0.0233 & 92.35 & 88.30 & 24.77 & 63.43 & 2.55 & 93.9 & 9.77 & 64.7\\
HuBERT \textsc{Base} \cite{Hsu2021HuBERTSS} &  94.68M  &  5.41 & 6.42 & 96.30 & 0.0736 & 98.34 & 88.53 & 25.20 & 64.92 & 2.58 & 93.9 & 9.36 & 65.8 \\
WavLM \textsc{Base} \cite{chen2022wavlm} & 94.70M &  \textbf{4.84} & \textbf{6.21} & 96.79 & \textbf{0.0870} & \textbf{98.63} & \textbf{89.38} & \textbf{22.86} & \textbf{65.94} & 2.58 & 94.0 & 10.37 & 66.6 \\
C-HuBERT \textsc{Base} & 96.00M &  6.14 & 7.38 & \textbf{96.92} & 0.0520 & 97.63 & 88.95 & 24.96 & 65.51 & \textbf{2.63} & \textbf{94.0} & \textbf{11.08} & 65.7 \\
\hline
wav2vec 2.0 \textsc{Large} \cite{Baevski2020} & 317.38M & 4.75 & 3.75 & 96.66 & 0.0489 & 95.28 & 87.11 & 27.31 & 65.64 & 2.52 & 94.0 & 10.02 & 65.5 \\
HuBERT \textsc{Large} \cite{Hsu2021HuBERTSS} & 316.61M & 3.53 & 3.62 & 95.29 & 0.0353 & 98.76 & 89.81 & 21.76 & 67.62 & 2.64 & 94.2 & 10.45 & 66.7 \\
WavLM \textsc{Large}${\dagger}$ \cite{chen2022wavlm} & 316.62M & \textbf{3.06} & \textbf{3.44} & \textbf{97.86} & \textbf{0.0886} & \textbf{99.31} & \textbf{92.21} & \textbf{18.36} & \textbf{70.62} & \textbf{2.70} & \textbf{94.5} & 11.19 & 68.4 \\
C-HuBERT \textsc{Large} & 318.95M & 3.78 & 4.02 & 96.79 & 0.0329 & 95.97 & 89.67 & 23.16 & 67.79  & 2.65 & 94.3 & \textbf{11.24} & 66.4 \\
\hline
\end{tabular}
\end{adjustbox}
\vspace{-10pt}
\end{table*}

Results are reported in Table~\ref{ms-asr-table}. First, we observe that both Cocktail HuBERT \textsc{Base} and \textsc{Large} significantly outperform the baselines, with \textsc{Large} reducing the WER by 69\% relative (24.9\% $\rightarrow$ 7.8\%). More importantly, there is a considerable gap between the HuBERT and Cocktail HuBERT performance (37.2\% vs 13.7\% for \textsc{Base}, and 35.2\% vs 7.8\% for \textsc{Large}), validating that the proposed masked pseudo source separation objective brings significant gain to the multi-speaker downstream task.

We further investigate the performance of our MS-ASR models on single speaker input by comparing with models fine-tuned for single-speaker ASR (Table \ref{ms-asr-single}). Overall, we see degradation in performance across all settings when using MS-ASR models to transcribe single-speaker utterances. However, compared to HuBERT, C-HuBERT models are more robust to variation in the number of speakers: the WER of \textsc{Large} increases by 1.2\% and 4.1\% on test-clean and test-other when fine-tuned on Libri2Mix instead of LS-10h, lower than the 5.1\%/9.1\% WER increases for HuBERT.

\begin{table}[t]
\vspace{5pt}
\caption{Diarization error rate (DER \%) results on LibriMix. DER reported for HuBERT and WavLM Base and Large models for 3 speakers and mixture of 2 and 3 speakers were obtained by us. C-HuBERT models are reported with $p_{mix} = 1.0$.}
\label{sd-librimix}
\vspace{-15pt}
\begin{center}
\begin{adjustbox}{width=0.85\columnwidth}
\begin{tabular}{lccc}
\hline
\multicolumn{1}{c}{\bf Method} & \multicolumn{1}{c}{\bf 2Mix} &\multicolumn{1}{c}{\bf 3Mix} &\multicolumn{1}{c}{\bf 2Mix + 3Mix} \\
 \hline
HuBERT \textsc{Base} \cite{Hsu2021HuBERTSS} & $5.88$ & $8.88$  & $9.04$  \\
WavLM \textsc{Base} \cite{chen2022wavlm} & $4.55$ & $7.13$ & $7.54$ \\
C-HuBERT \textsc{Base} & $2.77$ & $4.42$ & $3.95$ \\
\hline
HuBERT \textsc{Large} \cite{Hsu2021HuBERTSS} & $5.75$ & $7.84$ & $8.62$  \\
WavLM \textsc{Large} \cite{chen2022wavlm} & $3.24$ & $5.77$ & $5.62$ \\
C-HuBERT \textsc{Large} & $\textbf{2.65}$ & $\textbf{4.08}$ & $\textbf{3.86}$\\
\hline
\end{tabular}
\end{adjustbox}
\end{center}
\vspace{-10pt}
\end{table}

\subsection{Speech Diarization}

Table \ref{sd-librimix} shows results on speech diarization for two-, three-, and a mix of two- and three-speaker datasets. When compared against HuBERT and WavLM \textsc{Base} and \textsc{Large} models, the best DERs across the three settings are attained using C-HuBERT models. In fact, the C-HuBERT \textsc{Base} model outperforms the WavLM \textsc{Large} model on Libri2Mix, Libri3Mix, and Libri(2+3)Mix by $14\%$, $23\%$, and $30\%$ relative, respectively. These are impressive gains since the \textsc{Base} model is considerably smaller than the WavLM \textsc{Large} model and was pre-trained on fewer hours of data. Performance on all test sets are further improved when scaling to the \textsc{Large} model.

\subsection{SUPERB}
We compare C-HuBERT with several state-of-the-art SSL models on the SUPERB tasks (Table~\ref{superb}). C-HuBERT shows strong performance on speech enhancement and source separation, which are closer to the pre-training task of C-HuBERT. It lags behind other models on single-speaker tasks such as PR and ASR . However, the performance can be improved and the gap can be reduced by simply scaling C-HuBERT (on PR, 12\% PER reduction (1 - 5.41 / 6.14) between HuBERT and C-HuBERT for BASE and 7\% gap for LARGE).


\begin{table}[t]
\caption{Ablation study of mixing parameter: max number of utterances in the mixture $K$ and mixing probability $p_{mix}$.}
\label{ablation}
\vspace{-15pt}
\begin{center}
\begin{adjustbox}{width=\columnwidth}
\begin{tabular}{cc||cc|ccc|cc}
\hline
\multirow{2}{*}{$K$} & \multirow{2}{*}{$p_{mix}$} & \multicolumn{2}{c|}{SD} & \multicolumn{3}{c|}{MS-ASR} & \multicolumn{2}{c}{ASR} \\
& & 2Mix & 3Mix & dev-mix & d-c & d-o & d-c & d-o \\
 \hline
 & $0.2$ & $3.79$ & $6.70$ & $25.4$ & $\textbf{4.5}$ & $13.1$ & $4.0$ & $9.5$ \\ 
2 & $0.6$ & $3.48$ & $5.96$ & $13.3$ & $6.6$ & $14.9$ & $4.1$ & $9.8$ \\ 
 & $1.0$ & $3.35$ & $5.78$ & $16.1$ & $5.2$ & $14.4$ & $4.3$ & $10.1$ \\ 
\hline
 & $0.2$ & $3.65$ & $5.56$ & $26.3$ & $5.0$ & $\textbf{12.4}$ & $4.0$ & $\textbf{9.2}$ \\ 
3 & $0.6$ & $3.33$ & $5.14$ & $18.8$ & $7.1$ & $15.7$ & $4.2$ & $9.5$ \\ 
 & $1.0$ & $3.01$ & $4.88$ & $17.5$ & $5.1$ & $13.3$ & $4.3$ & $9.8$ \\ 
\hline
 & $0.2$ & $3.31$ & $5.30$ & $23.5$ & $9.5$ & $18.2$ & $\textbf{3.9}$ & $9.3$ \\ 
5 & $0.6$ & $2.97$ & $4.55$ & $\textbf{12.7}$ & $4.9$ & $13.0$ & $4.4$ & $10.3$ \\ 
& $1.0$ & $\textbf{2.77}$ & $\textbf{4.42}$ & $15.6$ & $5.5$ & $13.8$ & $4.6$ & $10.8$ \\ 
\hline
\end{tabular}
\end{adjustbox}
\end{center}
\vspace{-15pt}
\end{table}

\subsection{Ablation Studies}
We study the effect of mixing probability $p_{mix}$ and max number of speakers $K$ on speech diarization (SD), multi-speaker and single-speaker ASR (MS-ASR and ASR) with the C-HuBERT \textsc{Base} model (Table \ref{ablation}). On speech diarization, more aggressive mixing (larger $K$ and higher $p_{mix}$) leads to better results. The trend reverses on single-speaker speech recognition in general. Nevertheless, we observe that C-HuBERT outperforms HuBERT on single-speaker ASR for some configurations (e.g., $K=5$ and $p=0.2$ yields 3.9\%/9.3\% compared to 4.1\%/9.4\% from HuBERT in Table~\ref{ms-asr-table}).

We report results on both single- and multi-speaker test sets when fine-tuning C-HuBERT on multi-speaker data (columns below MS-ASR). Overall, $p_{mix}=0.2$ leads to the worst multi-speaker test results (23.5\% to 26.3\%), which are still better than those from HuBERT (35.8\%). The best multi-speaker test result is obtained with $K=5$ and $p_{mix}=0.6$. The results on single-speaker test sets (d-c, d-o) are interesting --- with $K=2$, single-speaker and multi-speaker WERs are negatively correlated, while with $K=5$ they are positively correlated. We believe the observation arises from the interaction of two factors: how mismatched pre-training and fine-tuning are, and how mismatched pre-training and testing are.

\section{Conclusion}
This paper presents Cocktail HuBERT, a large-scale pre-trained model with the objective of masked pseudo source separation. C-HuBERT extends the HuBERT framework to multiple speakers, enabling the models to outperform the state-of-the-art models on MS-ASR and SD tasks, while achieving competitive performance on other SUPERB single-speaker and multi-speaker tasks.





\bibliographystyle{IEEEbib}
\bibliography{refs}

\end{document}